# Background Subtraction on Depth Videos with Convolutional Neural Networks

Xueying Wang[1,2], Lei Liu[1] *, Guangli Li[1,3], Xiao Dong[1,2], Peng Zhao[1,2], Xiaobing Feng[1]

[1]State Key Laboratory of Computer Architecture, Institute of Computing Technology, Chinese Academy of Sciences, Beijing 100190, China

[2]School of Computer and Control Engineering, University of Chinese Academy of Sciences, Beijing 100049, China

[3]College of Computer Science and Technology, Jilin University, Changchun 130012, Jilin, China

{wangxueying, liulei, liguangli, dongxiao, zhaopeng, fxb}@ict.ac.cn

*Abstract*— Background subtraction is a significant component of computer vision systems. It is widely used in video surveillance, object tracking, anomaly detection, etc. A new data source for background subtraction appeared as the emergence of low-cost depth sensors like Microsoft Kinect, Asus Xtion PRO, etc. In this paper, we propose a background subtraction approach on depth videos, which is based on convolutional neural networks (CNNs), called BGSNet-D (BackGround Subtraction neural Networks for Depth video). The method can be used in color unavailable scenarios like poor lighting situations, and can also be applied to combine with existing RGB background subtraction methods. A preprocessing strategy is designed to reduce the influences incurred by noise from depth sensors. The experimental results on the SBM-RGBD dataset show that the proposed method outperforms existing methods on depth data, and even reaches the performance of the methods that use RGB-D data.

## I. INTRODUCTION

### A. Background Subtraction

Background subtraction is an important process for eliminating background and extracting concerned foreground in video analyzing and processing. It is widely used in computer vision systems, such as video surveillance, pedestrian and vehicle detection, hazard identification, etc. Traditional background subtraction methods are designed for RGB images. But they performs poorly in weak lighting and suddenly changed illumination situations. Low-cost depth sensors, for instance, TrueDepth cameras in Apple's iPhone X and Microsoft's Kinect, make background subtraction more challenging. The main difference between traditional data and depth data is the distance information got by depth sensors. A lot of issues about RGB background subtraction can be solved by the depth information, such as illumination changes, shadows, and color camouflage [1]. Depth sensors are insensitive to illumination changes and can get the scene information even if the environment is completely dark. Furthermore, shadows of objects will not be captured by the depth sensors, which makes sense in background subtraction task. Nevertheless, depth data are also limited by its fault, such as out of range points, noisy object boundaries and reflection [2].Thus, background subtraction methods based on depth data face severe challenges. Another challenge for depth data is that it is totally different from RGB value and the existing deep learning methods are only specific to color.

### B. Motivation

With the increasing emergence of surveillance devices, more and more public surveillance tasks have turned to relying on computer rather than human. However, most dangerous scenes are too dark to get enough color information, which brings challenges for crime detection and anti-terrorism.

Only depth data are used in this paper in order to achieve accurate depth background subtraction results without color information. Our method plays a crucial role in the dark scenes, where color information is hard to obtain. Our method can also improve the output images that are generated through multiple data sources.

With the rapid development of smart surveillance, computers have great importance in the production of modern industry. How to automate monitoring and controlling the industry sites has got much attention in recent years. While it cannot collect enough color information in dark room and some pixels are always black. It is vital to find an efficient and common way for monitoring in severe environments.

Traditional background methods usually construct complicated models and require a series of preprocessing and post-processing steps. While our method only contains a general preprocessing method that is also suitable for other methods and has no post-processing step, which achieves more accurate results.

### C. Overview and Contributions

In our work, a background subtraction method, which achieves more accurate results than all traditional methods based on depth data in SBM-RGBD, is proposed. A general preprocessing method is also designed to reduce the influences from edge noise and absent pixels. Firstly, we analyze depth images and find that zero numerical values which represent absent points have negative effects on the results. To solve these issues in processing depth data, we come up with an extended normalization method. Then we propose a new CNN architecture for depth data as illustrated in Fig. 1.

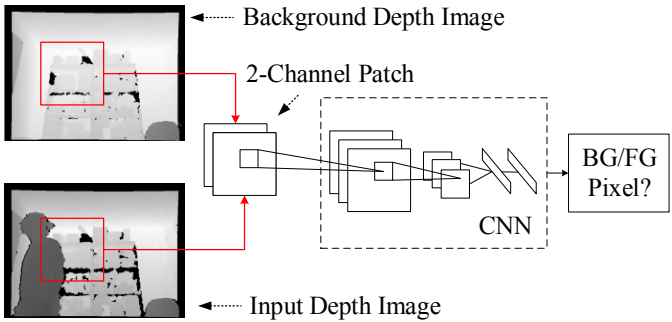

Fig. 1. Overview of BGSNet-D

As shown in Fig.1, 2-channel patches are extracted from the original input images which are composed of depth images and corresponding background images. Feature extraction and classification are accomplished by the convolutional neural networks. Pixel-wise classification results and foreground segmentation images from depth data can be obtained by computing the foreground probability. The proposed depth images background subtraction method is an independent component, which can be combined with methods based on RGB to improve the performance of background subtraction.

The contributions of this paper mainly include:

- *New CNN-Based Background Subtraction Method*: We propose a new background subtraction method, BGSNet-D, which can be used in the scenario where color images are unavailable, such as poor lighting situations.

- *Depth Data Preprocessing Strategy*: To process depth data and reduce the noise, a general depth data preprocessing strategy is proposed, which can be applied to all kinds of depth images processing tasks.

- *Experimental Study*: Experimental results show that BGSNet-D outperforms AvgM-D, GMM, Codebook and KDE on the SBM-RGBD dataset. Compared with the methods which take both color images and depth images as input, our method even achieves competing performance.

## II. RELATED WORK

### A. Background Subtraction with Depth Sensors

With the advent of low-cost deep sensors such as Microsoft Kinect, Asus Xtion PRO and Time-of-Flight (ToF) cameras, many researchers have studied how to use depth images to perform background subtraction. The existing background subtraction approaches for depth images are mostly traditional background modeling methods, which update background as video streams in, including Gaussian Mixture Model (GMM) [3], Codebook [4] and Kernel Density Estimation (KDE) [5], etc. These traditional methods were summarized and evaluated in [6].

GMM is a widely-used background subtraction method. It has successfully been used for generating background model. Massimo Camplani et al. [7] proposed a depth-color fusion background subtraction method. They analyzed the boundaries of depth images to improve the result, and used joint-bilateral filter (JBF) to remove the noise. In [8], they improved the results with JBF and update the model adaptively and iteratively. In [9], GMM background subtraction method was independently carried out on depth data and color features. Bayesian Network was used to predict the foreground object position. A region-based foreground predicting method was proposed in [10]. They introduced first public available RGB-D benchmark dataset in [11] and combined background subtraction algorithm with edge analysis, depth image classification and color image classification. In [12], a method combined GMM and Bayesian Network with probabilistic model was proposed. In [13], a noise model was proposed to preprocess depth data, where GMM was utilized to generate RGB and depth results respectively, and then combined the results together.

Codebook is another background modeling method. Enrique J. Fernandez-Danchez et al. [14] proposed a depth-extended method based on Codebook and used depth data to bias the segmentation results based on color. In [15], they further proposed another method based on Codebook algorithm which fuses range and color to remove most noise in disparity data by morphological reconstruction. Julian Murgia et al. [16] described a background subtraction method that utilizes colorimetric invariance to correct mistakes caused by illumination and combined depth information as external data sources.

Recently, KDE has been widely used for background subtraction. Daiela Giordano et al. [17] proposed a preprocessing method which filters or fills depth images by color information to reduce noise and deal with illumination changes. KDE was used to combine color and depth information for background subtraction after preprocessing. The recent dataset proposed by Gabrilel Moya et al. [18] is a new RGB-D dataset for foreground segmentation and generic scene modeling method was designed to analyze the different situations in depth images. Absent Depth Observations (ADO) are divided into two types: caused by scene's physical configuration and resulted from foreground objects edges. They classified ADOs by probability models and utilized KDE to eliminate background.

In RGBD-2017 [19], the workshop on background learning for detection and tracking from RGB-D Videos, there are some state-of-the-art works. Massimo De Gregorio, et al. [20] proposed an efficient background subtraction model based on weightless neural networks. The model is an extension of the previous work, CwisarDH, which designed for RGB videos. Lucia Maddalena, et al. [21] proposed a self-organizing neural background subtraction method based on two background models. The method combines the mask images of color and depth results. Sajid Javed, et al. [22] proposed a background subtraction method based on graph regularized Spatiotemporal RPCA. The method uses a semi-online algorithm and obtains the low-rank spatiotemporal information. Tsubasa Minematsu, et al. [23] used a modified visual background extractor, ViBe, as a background method. They combined the appearance and depth information using an energy function.

As mentioned above, most researches are based on the traditional background subtraction algorithm such as GMM, Codebook, KDE, etc. Depth data are usually preprocessed and integrated into RGB results with some rules. However, the background subtraction methods based on RGB-D data still have great potential to improve the accuracy of the results. In this paper, we use the advanced deep learning technology to focus on the accurate background subtraction of the depth only images. Our approach can also be used to combine other state-of-the-art background subtraction method, so as to achieve higher quality results.

### B. Deep Learning Methods for Background Subtraction

In 2012, AlexNet[24] won the first prize in the ImageNet Large Scale Visual Recognition Challenge (ILSVRC), which was far more accurate than other traditional visual algorithms, marked the rise of deep learning algorithm in the computer vision field. Subsequently, the convolutional neural network has been applied to the object detection [25], semantic segmentation [26] and other tasks, and has shown excellent performance, which fully demonstrates the capability of the deep neural network. Deep Neural Networks (DNNs) can extract features efficiently from images for classification and regression. Background subtraction method based on the deep learning is not the same as traditional ones, which predict by constructing background model as video streams in.

Marc Braham et al. [27] proposed a background subtraction method based on scene-specific convolutional neural networks. It is the first attempt to apply CNNs to the background subtraction task. They extracted the background images for scene-specific videos. After that, the neural network was trained with the image patches which are generated with the background images and input images. The label of the patch is obtained according to the corresponding pixel of ground truth image. They used trained CNNs to predict whether an image patch of central pixel is foreground or background. Mohammadreza Babaee et al. [28] proposed a general scenario background subtraction method that used a single CNN to estimate foreground images. They designed a new background extraction model and used spatial-median filtering as the post-processing. The network was trained with images from all videos and is able to handle various video scenes. Sakkos et al. [29] proposed a temporal-aware background subtraction method that used 3D convolutional neural networks. The method tracks temporal changes in video sequences and does not use any background model. Yi Wang et al. [30] proposed a CNN-based moving object segmentation method, which manually outlines certain amount annotated images with user intervention and trains cascaded CNN model to generate pixel accurate ground truth. The research is not a general purposed background subtraction method, but mainly focuses on the data annotation.

Existing background subtraction methods based on deep learning are all using a single data source, RGB images. Since depth data have different kinds of errors such as object boundary noise and invalid pixels [9], so it is hard to directly apply the prior method to depth image. In this paper, a deep data preprocessing method is proposed by analyzing the characteristics of the depth images. We use CNNs to learn features from depth data so as to achieve accurate segmentation outputs.

### III. PROPOSED METHOD

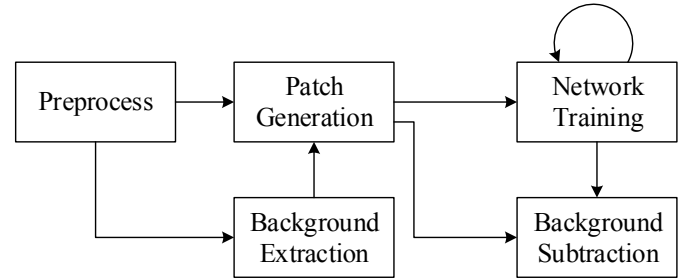

Fig. 2. Diagram of BGSNet-D

Our proposed method can be summarized along five steps: (1) preprocessing depth images; (2) extracting depth background patches; (3) generating normalized training data from background images and labeled images; (4) training the neural network; (5) predicting the foreground segmentation.

Note that only depth data are used in this method. Our primary goal is to achieve accurate depth background subtraction results without color information. There are two reasons for this. Firstly, it plays a crucial role in the dark scenes, where color information is hard to obtain. Secondly, it can also improve the outputs that generated by multiple data sources integrated background subtraction methods.

### A. Detpth Image Preprocess

Due to the nature of the depth sensors, there exists edge noise and pixel absence [9] in the captured data. Convolutional neural networks can extract features in color images, but cannot applied to depth images directly. For the sake of absent points, there are much zero numerical value in the images that does not mean zero distance, which may make existing CNNs incapable to learn and extract features.

The pixel value represents the distance between the object and the depth sensor, which is usually in millimeters. The minimum value is 0 which represents absent point and maximum value depends on the sensors that is different in most images.

In general RGB images, the value of a pixel distributes between 0 and 255. But in depth images, the value range varys in different scenes. Meanwhile, sensor errors and moving objects in the scenes make the minimum and maximum distance changes dynamically. In original depth images, the uncertain range makes it hard to extract features from patches when executing a convolution.

TABLE I.	NETWORK ARCHITECTURE

|  | **Conv1** | **Conv2** | **Conv3** | **FC1** | **FC2** | **FC3** |
|---|---|---|---|---|---|---|
| **Input Size** | 40×40×2 | 20×20×24 | 10×10×48 | 2400 | 1200 | 600 |
| **Filter Size** | 3×3 | 3×3 | 3×3 | - | - | - |
| **Conv Stride** | 1 | 1 | 1 | - | - | - |
| **Padding** | 1 | 1 | 1 | - | - | - |
| **Activation** | ReLU | ReLU | ReLU | Sigmoid | Sigmoid | Sigmoid |
| **Pooling Method** | max | Max | Max | - | - | - |
| **Pooling Size** | 2 | 2 | 2 | - | - | - |
| **Pooling Stride** | 2 | 2 | 2 | - | - | - |
| **Batch Norm/ size** | 2d/ 24 | 2d/48 | 2d/96 | 1d/1200 | 1d/600 | - |
| **Output Size** | 20×20×24 | 10×10×48 | 5×5×96 | 1200 | 600 | 1 |

In our experiments, the network is difficult to converge when using the original depth images for training. Normalization is a great idea to solve this problem. However, standard normalization may confuse the closest point value with the absent value. To solve the problems mentioned above, we proposed an extended min-max normalization method to preprocess the depth images:

$$x^* = \frac{x-(min-\alpha)}{(max-(min-\alpha))} \quad (1)$$

In the Equation (1), $x$ is the original depth value while $x^*$ is value after the specific normalization step, and $max$ represents the maximum distance in video, $min$ is minimum distance (except for failure points). $\alpha$ is a hyper-parameter, used to distinguish the missing points from effective points. By normalization, pixel values are scaled to the range between 0 and 1, in which minimum distance is mapped to $\alpha/(max-(min-\alpha))$, absent pixels equal 0 and maximum distance is 1. A big advantage of this is the value of minimum distance does not equate the value of the absent points. The processed images perform better on CNN training and feature extraction.

### B. Depth Backgorund Images Extraction

The input 2-channel patches are composed of average background images and the normalized depth images. Average background method is appropriate when the background pixel is visible for at least 50% of the time. In order to obtain high quality depth background image from video, an improved algorithm is proposed. To deal with sophisticated situations, the method in [31] was proposed. For depth images, we proposed a novel average background method to get rid of the noise in depth images. The improved background generation method is formulated as follows:

$$BG_{i,j} = \sum_{k=1}^{n} Img_{i,j}^k / (\sum_{k=1}^{n} valid(Img_{i,j}^k)) \quad (2)$$

$$valid(x) = \begin{cases} 1 & x \neq 0 \\ 0 & x = 0 \end{cases} \quad (3)$$

In the Equation (2) and Equation (3), $Img_{i,j}^k$ represents the depth value at pixel $(i, j)$ in $k^{th}$ input image, $BG_{i,j}$ displays corresponding depth value in background images and $n$ is the total frame number. The equation shows that the background image is the average of all depth input images except the invalid points. Invalid point effects are avoided and final background images are improved.

### C. Training Dataset Generation

A convolutional neural network requires data to learning feature so that we have to generate a large number of training samples. To learn the similarity between background and input, a 2-channel image patch is generated from background image and preprocessed input image, which both have the same central position. The label of the patch is given:

$$label(Patch_{i,j}) = \begin{cases} 1 & GT_{i,j} = FG \\ 0 & GT_{i,j} = BG \end{cases} \quad (4)$$

In the Equation (4), $Patch_{i,j}$ is the 2-channel patch of position $(i, j)$ and $GT_{i,j}$ is the ground truth. The label of the patch is 1 if the ground truth is foreground. In the ground truth images, there are some pixels that belong to the unknown motion which surround moving objects, neither the foreground nor the background. When generating training data, we will ignore these patches outside regions of interest.

### D. Network Architecture and Training

Our network architecture is shown in Table 1, which includes three building blocks. In each building block, a convolutional layer with 3×3 local receptive fields and a 1×1 stride is used. The rectified linear unit (ReLU) follows as the activation function in hidden layers. Batch normalization (BN) layer and pooling layer are after each ReLU layer. BN layer can both speed up the training procedure and improve the results. Experiments in following section are designed for further illustration. Max pooling layer is for parameter reduction and feature integration. Finally, all feature maps will input the Multilayer Perceptron (MLP) which contains two hidden layers. Sigmoid is used as activation function and the output layer only consists of a single unit.

By considering the problem as binary classification, we use a Binary Cross Entropy (BCE) function for training:

TABLE II. RESULTS ON SBM-RGBD DATASET

| Method | | Average Results | | | | | | |
|---|---|---|---|---|---|---|---|---|
| Data | Name | Recall | Specificity | FPR | FNR | PWC | Precision | F-Measure |
| Color + Depth (As a reference) | SCAD[23]* | **0.8847** | 0.9932 | 0.0068 | **0.0439** | 0.9088 | 0.8698 | **0.8757** |
| | RGBD-SOBS[21]* | 0.8391 | **0.9958** | **0.0042** | 0.0895 | 1.0828 | **0.8796** | 0.8557 |
| | SRPCA[22]* | 0.7786 | 0.9739 | 0.0261 | 0.1499 | 3.1911 | 0.7474 | 0.7472 |
| | cwisardH+[20]* | 0.7622 | 0.9817 | 0.0183 | 0.1664 | 2.8806 | 0.7556 | 0.7470 |
| Only Depth | BGSNet-D | 0.7827 | **0.9932** | **0.0068** | 0.1459 | **1.8872** | **0.8671** | **0.8138** |
| | AvgM-D[19]* | 0.7065 | 0.9869 | 0.0131 | 0.2221 | 2.8848 | 0.7498 | 0.7157 |
| | KDE[25] | **0.8095** | 0.9654 | 0.0346 | **0.1191** | 4.0925 | 0.6499 | 0.6981 |
| | Codebook[2] | 0.6962 | 0.9796 | 0.0204 | 0.2324 | 3.3686 | 0.7091 | 0.6745 |
| | GMM[1] | 0.6720 | 0.9619 | 0.0381 | 0.2566 | 5.2667 | 0.6050 | 0.5984 |

*Results of the SBM-RGBD Challenge @ RGBD2017

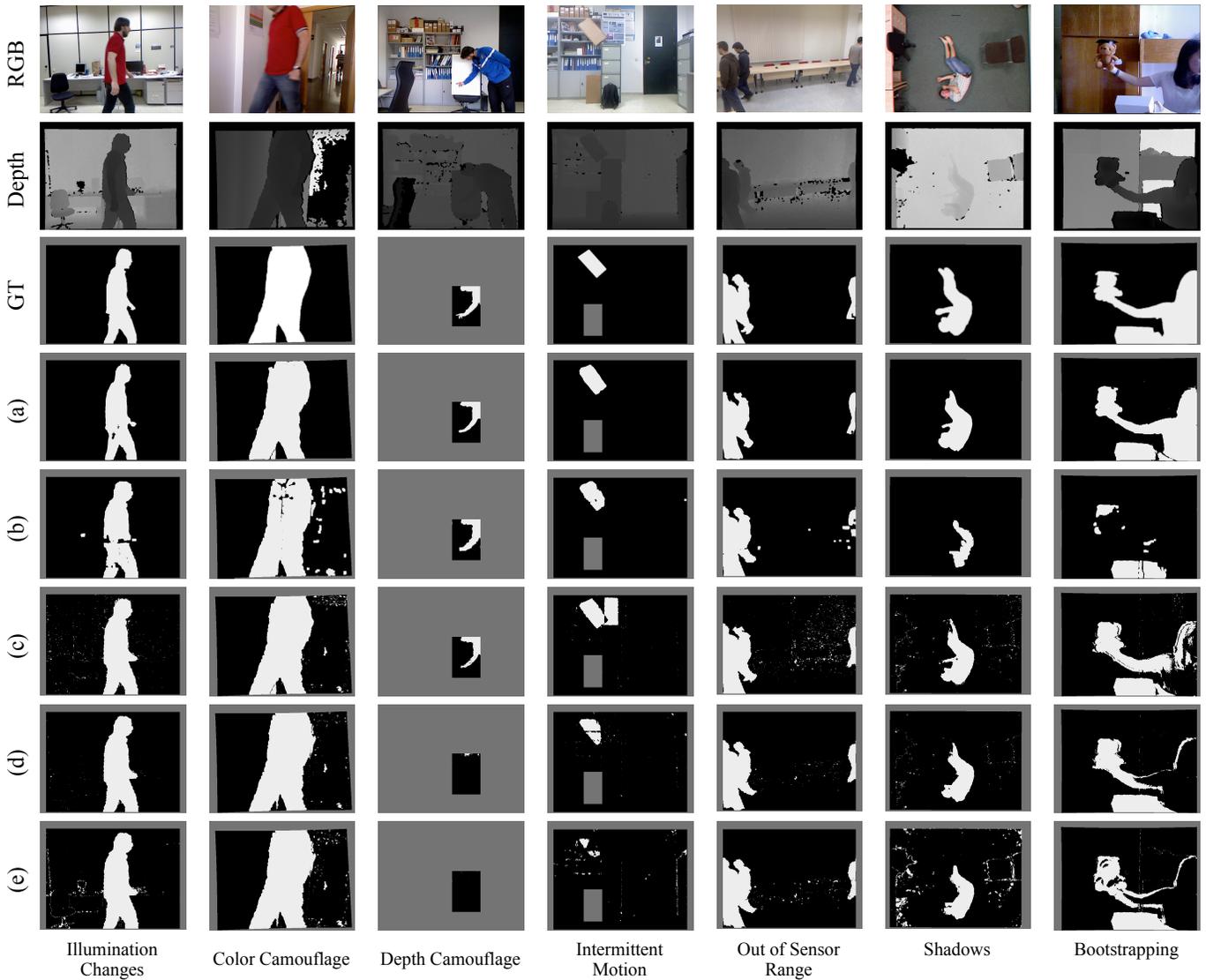

Fig. 3. Results on SBM-RGBD Dataset. (a)BGSNet-D (b)AvgM-D[19] (c)KDE[25] (d)Codebook[2] (e)GMM[1]

$$E = -\frac{1}{n}\sum_{i=1}^{n}(t_i \times log(y_i) + (1-t_i) \times log(1-y_i)) \quad (5)$$

where $y_i$ is the output of neural network and $t_i$ is the target value of background or foreground. RMSprop is used for back propagation while training. Learning rate is set to 0.001 while batchsize is 150.

## IV. EXPERIMENTAL RESULTS

### A. Experimental Settings

For fair comparison, we implement traditional methods with BGSLibrary 0 on depth data. We evaluate our proposed method BGSNet-D and these traditional methods on the SBM-RGBD [19] dataset. The experiments compare BGSNet-D with the state-of-the-art methods, including (1) AvgM-D [19], which only uses the depth data in RGBD2017[1], (2) GMM [3] used in [7-13], (3) Codebook [4] used in [14][16], (4) KDE [5] used in [17][18].

In order to compare and analyze the results of various background subtraction methods, the following evaluation metrics are used: recall, precision, specificity, false positive rate (FPR), false negative rate (FNR), percentage of wrong classifications (PWC) and F-measure, just the same as the metrics used in RGBD2017.

According to the rules in RGBD2017, the same parameters are used for processing all categories of videos. We implement the proposed method with PyTorch and set the size of patches T=40 and min-max normalization parameter $\alpha$ =10. All experiments are conducted on the machine with Intel Core i7-4770K CPU@ 3.50GHz and a TITAN X (Pascal) GPU which has 32G Memory.

### B. Experiments on SBM-RGBD Dataset

We report the experimental results of proposed method and other background subtraction methods in this section. The SBM-RGBD dataset is employed to test the performance of the methods, which provides a diverse set of labeled color images and depth images captured by Microsoft Kinect. The dataset consists of seven categories representative of typical indoor visual data captured in video surveillance and smart environment scenarios. It covers a wide range of background subtraction challenges, such as depth camouflage and out of sensor range.

In Table 2, the average results of every category on SBM-RGBD dataset are reported. As a reference, we also report the results of the background subtraction methods [[20]-[23]]. SCAD [23] won the RGBD2017 challenge, which used RGB-D data. Our proposed depth background subtraction method outperforms AvgM-D, KDE, Codebook and GMM, which reaches the performance of methods that use both RGB and depth data, even exceeds some of them.

Fig. 3 shows that our algorithm BGSNet-D outperforms other background subtraction methods which only use depth images. Each column in the figure represents the typical video frames in SBM-RGBD dataset. The first three lines are the RGB, depth and ground truth respectively, and (a)–(e) are the results of different background subtraction algorithms. Our approach shows accurate background subtraction results which have little noise, while the other methods have much more noise and need post-processing.

[1] http://rgbd2017.na.icar.cnr.it/SBM-RGBDdataset.html

The challenges for RGB background subtraction fall into three categories: *Illumination Changes*, *Color Camouflage* and *Shadows*. But for depth video sequences, they has few influence on background subtraction. Depth background subtraction methods can easily solve these problems which depth sensors cannot capture. *Bootstrapping* and *Intermittent Motion* are the situations which are too difficult to deal with in traditional background subtraction methods. For instance, (c) mistakes the sudden changed background regions for foreground, which generates the wrong results. Among the results of these categories, our outputs are much better than the other methods. *Depth Camouflage* category shows the videos in which foreground and background are very close to each other and hard to separate. Methods (e) and (d) are unable to generate acceptable result images. *Out of Sensor Range* contains the scenes that have unreachable regions which cannot be captured by depth sensors. Affected by these out of range points, results of (b) (c) (d) (e) have a lot of noise data. The BGSNet-D's F-measure metric, which represents the harmonic mean of precision and recall of results, is better in all the categories than the traditional methods which only use depth data. The result images in Fig. 3 demonstrate the effectiveness of the proposed approach and display advantages over other methods.

### C. Impact of Data Preprocessing

TABLE III. RESULTS ON PREPROCESSED DATA AND ORIGINAL DATA

| Method | Average Results (F-Measure) | |
|---|---|---|
| | *Original Data* | *Preprocessed Data* |
| BGSNet-D | 0.7811 | 0.8138 |
| KDE | 0.6030 | 0.6981 |
| Codebook | NaN | 0.6745 |
| GMM | NaN | 0.5984 |

The average results of various methods got from original data and preprocessed data are shown in Table 3. F-measure is not a number (NaN) in Codebook and GMM. The results are invalid due to mistaking all pixels for background, which means that recall is 0 and precision is NaN. After preprocessing, the accuracy of all kinds of methods is improved obviously, which validate the novel image preprocessing method proposed in this paper.

## CONCLUSION AND FUTURE WORK

In this paper, a background subtraction method BGSNet-D is proposed, which based on convolutional neural networks and processes data provided by depth sensors. BGSNet-D can be used in the scenarios where color information are unable to get. It extracts 2-channel patches from preprocessed images which are composed of input images and corresponding background images. To get the foreground probability of each pixel, feature extraction and classification are accomplished with CNNs. Experimental results on the SBM-RGBD dataset show that BGSNet-D outperforms all traditional methods mentioned in related work, including GMM, KDE, and Codebook. Our method can even compete the performance of background subtraction methods which uses RGB-D data. In the future, we will explore how to improve the performance by combining BGSNet-D with other RGB based methods.

ACKNOWLEDGMENT

This work is supported by National Key R&D Program of China under Grant No.2017YFB0202002, Science Fund for Creative Research Groups of the National Natural Science Foundation of China under Grant No.61521092 and the Key Program of National Natural Science Foundation of China under Grant Nos.61432018, 61332009, U1736208.